\documentclass[11pt]{article}
\usepackage[utf8]{inputenc}
\usepackage{graphicx,amsmath,amsfonts,amssymb,fullpage,xcolor,natbib}

\usepackage[T1]{fontenc}

\title{\textbf{Subjective functions}}
\author{Samuel J. Gershman\\
Department of Psychology and Center for Brain Science\\
Kempner Institute for the Study of Natural and Artificial Intelligence\\
Harvard University}

\date{}

\begin{document}

\maketitle

\begin{abstract}
    Where do objective functions come from? How do we select what goals to pursue? Human intelligence is adept at synthesizing new objective functions on the fly. How does this work, and can we endow artificial systems with the same ability? This paper proposes an approach to answering these questions, starting with the concept of a subjective function, a higher-order objective function that is endogenous to the agent (i.e., defined with respect to the agent's features, rather than an external task). Expected prediction error is studied as a concrete example of a subjective function. This proposal has many connections to ideas in psychology, neuroscience, and machine learning.
\end{abstract}

\section{Introduction}

Objective functions are central to all learning systems (both natural and artificial). The way we distinguish learning from other kinds of dynamics is the fact that learning produces (at least in expectation or asymptotically) an improvement in performance as measured by an objective function.\footnote{For example, passive wear and tear degrades the function of living organisms and robots over time, but this is not learning, because it cannot be understood in terms of performance improvement over time.} Many different objective functions have been proposed, and it's not clear that all intelligence can be subsumed by a single ``ultimate'' objective, such as reproductive fitness.\footnote{Even the reasonable argument that all forms of biological intelligence arose from natural selection is in practice not very helpful for elucidating the underlying principles that give rise to intelligent behavior.} Perhaps the problem is that the quest for a single objective function is misguided. An important characteristic of human-like intelligence may be \emph{the ability to synthesize objective functions}.


To gain more insight into this peculiar ability, we need a framework for deriving objective functions from the agent itself. We will call this mapping (formalized in the next section) a \emph{subjective function}. Because it is endogenous to an individual agent, it cannot be conceptualized as a universal objective function---it depends intrinsically on the agent's internal state. 

To understand what this means, consider a typical way to define an objective function: stipulate some reward or supervision signal, then score an agent based on how well it maximizes expected reward or minimizes expected error. These signals are exogenous to the agent in the sense that their definitions do not depend on any feature of the agent; they can be applied uniformly to any agent. In contrast, a subjective function is endogenous to the agent in the sense that the definition of the signal that the agent is optimizing depends on features of the agent. Stated another way: subjective functions are a way of modeling how agents choose what goals to pursue, rather than (in the standard reinforcement learning framework) pursuing goals that are given to them.

This paper is organized as follows. The next section introduces a general theoretical framework for understanding the relationship between objective (value) and subjective (meta-value)  functions. We describe a set of criteria for defining a ``good'' subjective function, and then formalize a specific subjective function (expected prediction error) satisfying these criteria. We show how this subjective function can be used to design an agent capable of open-ended learning. We then discuss how it connects to observations from psychology and neuroscience, as well as related ideas in machine learning.

\section{Theory of subjective functions}

\subsection{Goal-conditioned reinforcement learning as a meta-MDP}

Following a standard reinforcement learning (RL) setup, we model a task as a Markov decision process (MDP) consisting of the following components:
\begin{itemize}
    \item A state space $\mathcal{S}$.
    \item An action space $\mathcal{A}$.
    \item A transition distribution $T(s'|s,a)$, where $s,s'\in \mathcal{S}$ and $a \in \mathcal{A}$. This is the probability of transitioning to state $s'$ after taking action $a$ in state $s$.
    \item The agent chooses actions according to a policy $\pi(a|s)$, the probability of choosing action $a$ in state $s$.
    \item A reward function $R(s)$.
\end{itemize}
Importantly, we do not assume a fixed reward function. Instead, we allow the agent to select its own reward function. For concreteness, we will study the case where the reward function is parametrized by a specific goal state $g \in \mathcal{S}$:
\begin{align}
    R_g(s) = \mathbb{I}[s = g].
    \label{eq:reward}
\end{align}
Thus, the reward is 1 only when the agent has reached the goal state. Parametrizing the reward function in terms of dynamic goals is known as \emph{goal-conditioned reinforcement learning} \citep{liu2022goal}. The goal-based framework is appealingly simple and applicable to many environments that are natural for humans. It's straightforward to extend this setup (e.g., to reward functions that are linear in some feature space), where $g$ is interpreted as a set of \emph{reward parameters}.

A standard objective function in RL is the expected discounted future reward, or \emph{value}:
\begin{align}
    V^\pi_g(s) &= \mathbb{E}\left[\sum_{t=0}^\infty \gamma^t R_g(s_t) \bigg| s_0 = s, \pi, g\right] \nonumber \\
    &= R_g(s) + \gamma \sum_{a} \pi(a|s) \sum_{s'} T(s'|s,a) V^\pi_g(s'),
\end{align}
where $t$ indexes time and $\gamma \in [0,1)$ is a discount factor governing how the agent values long-term reward. The second equality is the Bellman equation.

Temporal difference (TD) learning is a classical technique for estimating the value function \citep{sutton1988learning}. Conditional on goal $g$ and policy $\pi$, a value function approximation $\hat{V}^\pi_g$ is learned by optimizing the expected squared TD error objective function, $\mathbb{E}[\delta_t^2]$, where
\begin{align}
    \delta_t = R_g(s_t) + \gamma \hat{V}^\pi_g(s_{t+1}) - \hat{V}^\pi_g(s_t)
\end{align}
is the TD error, also known as the \emph{reward prediction error} because it quantifies the discrepancy between received and predicted rewards. TD learning is typically applied online, using a stochastic approximation of the gradient.

To model goal selection, we nest tasks (base-level MDPs indexed by goals) within a meta-MDP with state space $\mathcal{M}$, where each $m \in \mathcal{M}$ corresponds to a tuple $m = (g, \omega)$. The parameter $\omega$ represents the \emph{agent state}---i.e., the internal aspects of the agent that are used to specify the goal. Actions, generated by a meta-policy $\tilde{\pi}(g|m)$, correspond to goal choices, leading to transitions in the agent state through its interactions with the task MDP. We can now define a subjective function more precisely: it is the value function of the meta-MDP. Its subjectivity derives from its dependence on the agent state.

The meta-MDP formalism\footnote{See also \citet{Paul2026} for a related approach to MDP selection within a meta-MDP framework (more precisely, they define a meta decision problem for a class of extended partially observable MDPs). That work is focused on the problem of ``centering'' the agent within a representation of the environment in situations where the agent's self identity is ambiguous.} departs from a basic premise of most approaches to RL---that the reward function depends only on the environment state. Note that even apparently agent-dependent properties can be accommodated within the standard RL framework by ``externalizing'' them (pushing them into the environment state). For example, hunger is ostensibly an internal property of an agent, but we can externalize it by shifting the boundaries of the agent and the environment \citep[see][]{niv2006normative,keramati2014homeostatic,juechems2019does}. In fact, any meta-MDP can be expressed as a ``flat'' MDP with an augmented state space. So what is the advantage of the meta-MDP formalism?

One computational advantage is that it allows the agent to optimize its policy in a much smaller state space, once the goal has been fixed. Specifically, the meta-MDP lends itself to a bilevel optimization scheme in which the agent alternates between goal selection and goal pursuit. A potential disadvantage of this scheme is that the agent may perseverate in pursuing inauspicious goals. Indeed, such perseveration is characteristic of human goal pursuit \citep{shah2002forgetting,cheng2023intention,holton2024goal,aenugu2025building}. Nonetheless, perseveration may actually be a useful asset in conditions where many goals compete for attention, computational resources are scarce, or rewards are sparse \citep{holton2025adaptive,prystawski2022resource}. Later in the paper, we will build the case that the meta-MDP formalism is not only computationally advantageous but also psychologically plausible.

\subsection{What makes a ``good'' subjective function?}

Within the meta-MDP framework, any choice of subjective function is as ``good'' as any other, in the narrow technical sense. However, there are several intuitive criteria for preferring some subjective functions over others.

First, we would like agents that achieve broad coverage of the goal space. While agents that specialize in one or a few goals might do perfectly fine in their ecological niches, it is thought that general intelligence requires broad coverage \citep{lake2017building,chollet2019measure}. Relatedly, we would like subjective functions that don't create pathological loops, where the agent cycles between a small set of goals.

Second, we would like goal coverage to expand efficiently. This means prioritizing achievable goals given the current agent state, and briskly moving on to the next goal once the current one is achieved.

Third, parsimony favors ``compatible'' subjective functions that don't require significant additional machinery to compute beyond what is already available for dealing with the base-level MDP. In other words, we would like agents that can reuse computations at the meta-level that they are already using at the base level.

Fourth, we want subjective functions that are hard to game by self-deception. Once we allow agents to choose their own reward functions, what prevents agents from entering perpetual bliss where every state is infinitely rewarding? Presumably such blissful lives are cut short by negative fitness; if you enjoy being someone else's meal or falling off a cliff, you won't survive long.

Fifth, we want subjective functions that are robust to thorny design choices about state features (e.g., whether hunger is included in the state representation).

In the next section, we introduce a subjective function that satisfies these criteria.

\section{Expected prediction error}

A meta-level policy $\tilde{\pi}(g|m)$ that optimizes value would repeatedly select the nearest achievable goal, violating the coverage criterion described above. What's needed is a ``quenching'' mechanism for completed goals, while still incentivizing the agent to pursue achievable goals. One way to do this is to pursue goals that yield better-than-expected reward. Positive prediction error provides a natural goal gradient, indicating the path that will bring the agent closer to the goal. Once this path is discovered, the prediction error vanishes. Importantly, the meta-level policy also needs to be predictive, anticipating which goals will lead to positive prediction error. This brings us to the concept of \emph{expected prediction error} (EPE), the expected discounted sum of future prediction errors ($\delta_t$):
\begin{align}
    U^\pi_g(s) = \mathbb{E}\left[\sum_{t=0}^\infty \gamma^t \delta_t \bigg| s_0 = s, \pi, g\right]. \label{eq:goalprogress}
\end{align}
Notice that we have simply replaced the rewards in the standard value function with TD errors; in other words, the value function of the meta-level MDP (with agent state $\omega = s$) is the expected TD error for a given base-level state. The EPE measures a form of goal progress, because $\delta_t \approx \dot{V}^\pi_g$ prior to goal attainment (i.e., the prediction error approximates the rate of change in the goal-dependent value).

Eq. \ref{eq:goalprogress} is a telescoping series (intermediate terms cancel out), allowing us to express it as follows:
\begin{align}
    U^\pi_g(s) = V^\pi_g(s) - \hat{V}^\pi_g(s), \label{eq:telescope}
\end{align}
where we have assumed that $\hat{V}^\pi_g$ is bounded so that $\lim_{T \rightarrow \infty} \gamma^T \hat{V}^\pi_g(s_T) = 0$. Because the TD error is used to update value estimates, it is also necessary to assume that $\hat{V}^\pi_g$ is frozen (or slowly changing) when computing $\delta_t$. This is similar to the logic of target networks in deep RL.

Eq. \ref{eq:telescope} shows that optimizing EPE is equivalent to maximizing value up to a constant (the frozen value estimate); this means that an agent maximizing EPE will try to improve its expected future rewards. However, if the value estimate is perfect for a state, EPE is 0; this means that the agent will not try to reach high-value states that it knows are high value. EPE is subjective in the sense that the same value function can lead to different utilities depending on the agent's value estimate.

The essence of EPE is that agents are attracted to positive surprise. This still requires agents to figure out how to get to the goal, but once they arrive, the goal is quenched. For example, once you reach the end of a maze, it's no longer of any interest to stick around and relive the victory, but that's precisely what a value-maximizing agent would do if the end point is rewarding and the game doesn't terminate. For the same reason, it's of no interest to repeatedly retrace the path to victory, but that's precisely what a value-maximizing agent would do if given the opportunity.

A slightly more complicated model combines value and EPE, to accommodate the fact that even in the limit of perfectly learned values (where EPE is 0) agents may still prefer policies with higher values:
\begin{align}
    \alpha V^\pi_g(s) + (1-\alpha) U^\pi_g(s) = V^\pi_g(s) - (1-\alpha) \hat{V}^\pi_g(s),
\end{align}
where $\alpha \in [0,1]$ is a weighting parameter ($\alpha=0$ recovers the EPE model; $\alpha=1$ recovers the value model). In the limit where $\hat{V}^\pi_g = V^\pi_g$, the model reduces to $\alpha V^\pi_g$ (i.e., a dampened version of the value model). When $\alpha$ is close to 0, optimal policies will pursue error maximization until learning has eliminated most sources of error, at which point policies will pursue value maximization.

This generalized version of EPE is suitable as an objective function for the base-level MDP, because it corresponds to classical value maximization for a fixed goal. It is also suitable as the subjective function (possibly with a different setting of $\alpha$) for the meta-level MDP, because: (i) it encourages coverage by quenching  completed goals; (ii) it accomplishes efficient expansion of coverage by selecting goals that are neither too easy nor too hard (both of which will yield 0 or negative prediction error); (iii) it is compatible in the sense that the same function can be reused at both the base and meta-levels; and (iv) it is hard to game by self-deception because making everything rewarding would drive the EPE to 0, whereas inflating errors by (for example) slowing down learning will reduce value.\footnote{From a psychological perspective, it's possible that humans exhibit a spectrum of $\alpha$ values which determine individual susceptibility to reward hacking / self-deception.} The agent will continually strive to select new goals that it doesn't yet know how to achieve, leading to truly open-ended learning. Thus, the EPE satisfies the criteria for subjective functions laid out in the previous section. Comparisons with other possible subjective functions are considered below.

\section{Connections to psychology and neuroscience}

\subsection*{Hedonic adaptation}

You might think that you're happiest when good things happen, but evidence suggests that people become rapidly desensitized to rewarding stimuli, a phenomenon known as \emph{hedonic adaptation} \citep{frederick1999hedonic}. Some examples:
\begin{itemize}
    \item Lottery winners are not in general happier than other people, and in fact take less pleasure in mundane events \citep{brickman1978lottery}.
    \item Repeatedly consuming an initially desirable food reduces its pleasantness and subsequent consumption \citep{rolls1981sensory}.
    \item Desensitization to certain pleasurable activities (e.g., drug-taking) may drive the formation of addictive behaviors as a form of compensation \citep{koob1996drug}.
\end{itemize}
These observations are consistent with the idea that goal attainment quenches incentive (formalized by reduction of EPE).

A quantitative analysis of momentary subjective well-being indicated that well-being judgments are strongly predicted by the history of recent prediction errors in a gambling task \citep{rutledge2014computational}. This supports the idea that prediction errors themselves are subjectively valuable.

\subsection*{Preference for increasing reward}

When given a choice between sequences of outcomes, people usually prefer sequences of increasing expected reward \citep{loewenstein1993preferences}. For example, people prefer increasing sequences of payments \citep{loewenstein1991workers}, even if this results in lower total income \citep{hsee1991relative}. Similarly, people prefer sequences of decreasing discomfort to sequences of increasing discomfort \citep{varey1992experiences,chapman2000preferences}. Reports of satisfaction and positive mood are also higher for increasing sequences \citep{hsee1991velocity,lawrence2002velocity}. This is puzzling from the perspective of standard economic theory, because it seems to imply a negative discount rate ($\gamma < 0$)---i.e., a preference for smaller rewards sooner. However, it makes more sense from the EPE perspective: prior to goal attainment, prediction errors are approximately the temporal derivative of estimated value. Thus, maximizing expected prediction error leads to preferences for increasing expected reward over time.

\subsection*{Information avoidance and demand}

In states where value estimates tend to be optimistic ($\hat{V}^\pi_g > V^\pi_g$), EPE is negative, and therefore agents will tend to avoid policies such as information gathering that might reduce the optimism gap. Indeed, optimism bias is widespread \citep{sharot2011optimism}, and may be a driver of information avoidance \citep{golman2017information}. For example, \citet{eil2011good} found that people tend to avoid information about personal attributes like attractiveness or intelligence when they receive a hint that the information may lower expectations. Similarly, people at risk for Huntington disease tend to both underestimate their risk and avoid genetic testing \citep{oster2013optimal}.

At first glance, these findings seem opposed to a different set of findings indicating a preference for early information revelation, even when that information is not instrumental (i.e., it cannot change future outcomes). \citet{kendall1974preference} gave pigeons the choice between a deterministic option (reward was always delivered, preceded by a white light) and a random option (reward was delivered 50\% of the time, preceded by a green light when reward would be delivered, or by a red light when reward would not be delivered). Pigeons preferred the random option, even though it gave them half as much reward. Critically, they only preferred the random option when the lights were predictive; they strongly preferred the deterministic option when the lights were uncorrelated with reward delivery.

One way to understand the pigeons' apparently suboptimal choices starts from the hypothesis that their value estimates are pessimistic  ($\hat{V}^\pi_g < V^\pi_g$) and hence EPE is positive. This could arise from the delay between the light and reward, which introduces noise into magnitude estimation; Bayesian filtering of this noise regularizes the estimate towards the prior \citep{gabaix2017myopia,gershman2020rationally}. If the prior expectation is less than $V^\pi_g$, the result is underestimation. This account is consistent with the observation that suboptimal choice prevails primarily when the delay is long \citep{dunn2024suboptimal}, which is also when noise should be larger and regularization stronger. Under the pessimism hypothesis, agents should demand information which might reduce the pessimism gap.

Another source of data relevant to this hypothesis comes from neurophysiology. Dopamine neurons, which are thought to report prediction errors \citep{gershman2024explaining}, increase their activity in response to informative cues, and decrease their response to uninformative cues \citep{bromberg2009midbrain}. Moreover, the difference between the responses to informative vs. uninformative cues predicted the animal's preference for informative cues. Thus, it is plausible that this preference is driven by expected prediction errors, though the study does not establish causality.

In summary, the EPE model predicts information avoidance when values are overestimated ($\hat{V}^\pi_g > V^\pi_g$) and information demand when values are underestimated ($\hat{V}^\pi_g < V^\pi_g$). These predictions are broadly consistent with empirical data.

Another way to think about these observations is in terms of temporal discounting applied to prediction errors rather than rewards. When prediction errors are expected to be negative, agents should seek to defer them as long as possible. When prediction errors are expected to be positive, agents should seek to receive them as soon as possible.\footnote{\citet{iigaya2016modulation} and \citet{zhu2017information} have developed related, but somewhat different, accounts of information demand based on the idea of maximizing prediction errors.}

\subsection*{Conditional rationality in goal pursuit}

The principle of rational action states that agents will adopt the most efficient policy for achieving a goal (e.g., they will take the shortest available path to a goal location). In other words, agents should maximize value. As stated earlier, maximizing EPE is equivalent to maximizing value (as long as the value estimate is fixed or changing sufficiently slowly). Critically, the value function itself is determined by the agent's goal selection, which optimizes the same subjective function. This results in what has been called \emph{conditional rationality} \citep{chu2024praise}: efficient pursuit of subjective goals.\footnote{The concept of conditional rationality has deep roots in moral philosophy. In a famous passage of his \emph{Treatise of Human Nature} \citep{hume2000treatise}, the philosopher David Hume concluded: ``Reason is, and ought only to be the slave of the passions, and can never pretend to any other office than to serve and obey them.''}

The paradigmatic example of conditional rationality is children's pretend play. To a large extent, this form of play follows realistic rules/constraints---up to a point \citep{harris2021early}. For example, when 2-year-olds watch as pretend toothpaste is squirted onto one of two toy pigs, they correctly clean the `dirty' pig \citep{harris1993young}. Clearly the pretend squirting action violates the real-world constraint that the toothpaste should be visible, but nonetheless children follow an efficient plan \emph{conditional} on the premise that toothpaste has been squirted on a particular pig.

Experiments with 4- and 5-year-olds take this idea one step further \citep{chu2023not}. In one experiment, children were brought into a room with pencils attached to the wall; some pencils were `low-cost' (could be reached easily), whereas others were `high-cost' (required jumping). When instructed to retrieve the pencil (an exogenously specified goal), most children followed the principle of rational action, taking the low-cost action. In contrast, children instructed to play (``Could you play over there? Maybe you could play a game to get the pencil.'') preferentially took the high-cost action---jumping is inefficient (from the perspective of pencil collection) but fun!

Similar results were obtained with a sticker collection task: a box of stickers was placed on the floor at the end of a spiral constructed out of tape and colorful dots. When instructed to retrieve the stickers, all children walked straight to the box, ignoring the spiral. When instructed to play, most children walked along the spiral. These studies tell us that an important part of children's play is selecting goals that may look quite different from the goals selected exogenously by adults. Nonetheless, children pursue these goals efficiently: jumping and following the spiral are ``efficient'' plans in pursuit of endogenously selected goals.

One way to think about these findings is that when children are encouraged to choose their own goals, they tend to choose ones that produce less predictable but more informative outcomes. For example, jumping or walking along a spiral will produce variability that informs children about their competence at these skills. This is precisely what maximizing EPE achieves: future prediction errors that stimulate learning.\footnote{See also \citet{Andersen2023} for a related analysis of play in terms of maximizing future learning opportunities.}

Lest you think this applies only to children, consider some examples from the Guinness Book of World Records:
\begin{itemize}
    \item In 2016, the largest DNA helix composed of humans (4000 participants) was assembled on a beach in Varna, Bulgaria.
    \item At the 2009 National Window Cleaning Competition in Blackpool, UK, Terry Burrows became the fastest window cleaner in history by cleaning three standard office windows in 9.14 seconds.
    \item In 2014, Bruce Masters achieved the record of ``Most Pubs Visited'' (46,495).
\end{itemize}
These goals are essentially arbitrary, in the sense that there is no instrumental logic dictating which goal to pursue. But once selected, people pursue them doggedly. The Guinness Book of World Records is a sourcebook of conditional rationality taken to its extremes.

A less fanciful but more systematic study of conditional rationality in adults was undertaken by \citet{cushman2015habitual}. Using a 3-step sequential decision problem, they showed that people tend to follow policies that bring them efficiently to a goal which had been previously rewarding in the past, even when this results in a globally suboptimal policy. This behavior was consistent with a model that learned goal values by TD learning, but could also be consistent with a goal-selection model based on EPE.

Conditional rationality can give rise to pathological behaviors. Although it is often thought that addiction reflects compulsive habit formation that eventually supersedes goal-directed control of behavior \citep[e.g.,][]{everitt2005neural}, this view is incompatible with observations of sophisticated goal pursuit in addicts \citep{simon2012dual,hogarth2020addiction}. For example, people seeking prescription drugs sometimes fabricate or tamper with electronic medical records. People who engage in `doctor-shopping' behavior (moving between providers until they receive a prescription) are highly effective at reaching their goals \citep{schneberk2020opioid}. A study of heroin abusers \citep{johnson1985taking} documented that daily users consume about $\$36$ worth of heroin per day. To pay this cost, they engage in structured economic activities (often in the heroin industry itself). Presumably these activities require goal-directed planning. Thus, our ability to efficiently pursue goals may not always be compromised in drug addiction; rather, it may become hijacked by drug-directed goal selection.

\subsection*{Task selection and hierarchical reinforcement learning}

Experiments suggest that people prefer tasks (equivalent to goals in this setting) that lead to performance improvements \citep{ten2021humans,poli2022contributions}; Similar results have been reported in 4-year-olds \citep{poli2025exploration}. This means that people select tasks that are not too easy and not too hard, depending on their current performance. Intuitively, if a task is too easy or too hard, goal progress (measured by the EPE) will be close to 0. This principle is closely related to the concepts of \emph{learning progress} and \emph{competence progress} in machine learning, discussed below.

In some cases, task selection can be modeled as a form of hierarchical RL, such as in the \emph{options framework} \citep{sutton1999between}. An option is a temporally abstract action; once selected, it invokes a lower-level policy until the option is terminated. For example, foraging involves the selection of a patch (the abstract action), followed by harvesting of the patch with lower-level motor actions like approaching a tree, picking fruit, etc. \citep{alejandro2024hierarchical}. Technically, options define a \emph{semi-Markov} MDP, which extends the classical MDP to allow for random dwell times in each state. These dwell times correspond to the option durations. This is fundamentally different from the meta-MDP framework, where task or goal selection \emph{defines} the base-level MDP. One way to appreciate this distinction is to notice that the options framework requires a single reward function regardless of the option choice. There's no sense in which the options framework allows the reward function to change, unlike the meta-MDP framework. This distinction is important for conceptualizing how reward functions can change radically after transformative experiences \citep{paul2014transformative,Paul2026}. The decision to have a baby, for example, is more than an abstraction action like choosing a patch in which to forage; it involves radical change in what a person finds rewarding in life.

\section{Connections to machine learning}

\subsection*{Prediction error as intrinsic reward}

Intrinsic rewards are typically used in RL systems as additional terms added to extrinsic rewards in order to shape behavior in particular ways \citep[e.g., to encourage exploration and curiosity, or to provide an inductive bias for policy learning;][]{poli2024curiosity,gruaz2025merits}. Subjective functions operate at the meta-level, determining the reward function used by the RL system at the base level. These two approaches are not incompatible; a subjective function could in principle be used to define an intrinsic reward at the base level. In this section, we discuss how prediction errors have been used to define intrinsic rewards in the RL literature, and how this is related to the EPE subjective function.

\citet{simmons2020reward} trained two parallel function approximators, which differed only in the definition of reward: an ``exploitation'' approximator using the standard (extrinsic) reward, and an ``exploration'' approximator using the absolute value of the TD error (intrinsic reward). The exploration policy controls actions during training, whereas the exploitation policy controls actions at test time. This produces high-error training examples that encourage exploration, while the exploitation approximator learns the optimal values off-policy \citep[see][for a closely related approach]{griesbach2025learning}. \citet{gehring2013smart} also used absolute TD error as an intrinsic reward (what they called \emph{controllability}), adding it as a bonus to the value function during action selection. The variance of TD errors has also been used as an intrinsic reward \citep{flennerhag2020temporal}. All of these approaches share the aim of encouraging exploration towards error-generating parts of the state space.

Most closely related to the ideas here is the \emph{Positive Error Bias} algorithm \citep{parker2025biasing}, which defines a softmax policy over an estimate of the expected TD errors for each action. They also studied a version of the model where the expected TD error estimator is used only to drive feature learning, while actions are controlled by a value estimator based on the same features.

A key advantage of using signed TD errors (as in the Positive Error Bias algorithm and the EPE subjective function) compared to unsigned (e.g., absolute) TD errors is that the agent is diverted away from states associated with negative TD errors. Agents that pursue positive surprise will (as we've shown) provably maximize value relative to their estimates. In contrast, agents that pursue both positive and negative surprise may end up spending significant time in aversive states.

\subsection*{Learning progress and competence progress}

A related line of work in model-based systems uses the \emph{unsigned} error of model predictions to guide exploration and goal selection \citep[e.g.,][]{schmidhuber1991possibility,schmidhuber2010formal,oudeyer2007intrinsic,pathak2017curiosity,molinaro2024latent}. For example, \citet{oudeyer2007intrinsic} developed an agent that seeks out states where squared error is expected to decrease---i.e., states with high \emph{learning progress}. An important insight from this work is that improvement is a better intrinsic reward than the current performance level, because the latter leads agents to getting stuck in highly unpredictable regions of the state space. The main challenge for this kind of approach is that error in sensory space can be very noisy. \citet{pathak2017curiosity} try to mitigate this problem by predicting actions instead.

More closely related to the proposal here is the idea of using unsigned TD errors as an intrinsic reward \citep[reviewed in][]{baldassarre2012deciding}. For example, \citet{schembri2007evolution} developed an agent consisting of several ``experts'' that learn action policies and a ``selector'' that decides which expert is in control at any given time. During a ``childhood'' (exploratory) phase, the selector is trained using the TD error of the selected expert as reward. In this way, it selects experts whose competence is expected to improve, and thereby improves the competence of the system as a whole. \citet{stout2010competence} study a similar idea, framed in terms of temporally extended skills rather than experts. Importantly, these approaches avoid the issue of noisy errors in high-dimensional sensory space.

Using the unsigned TD error as an intrinsic reward is one version of a more general family of algorithms that use \emph{competence progress}---performance improvement over the course of learning---to guide exploration and goal selection \citep{oudeyer2007intrinsic,baranes2010maturationally,colas19,colas2022autotelic}.

Generally speaking, using unsigned prediction errors as the subjective function has the property that agents will pursue goals that may be \emph{less} rewarding than expected (i.e., negative prediction errors). In contrast, the subjective function based on EPE always drives the agent towards goals that are expected to produce positive prediction errors.

\subsection*{Generalized advantage estimation}

The EPE is formally related to an important algorithmic concept widely employed in the reinforcement learning literature. The \emph{advantage function} $A^\pi(s,a)$ is defined as the difference between the state-action value function and the state value function:
\begin{align}
    A^\pi(s,a) = Q^\pi(s,a) - V^\pi(s) = \mathbb{E}[\delta|s,a, \pi].
\end{align}
The second equality shows that the advantage function is the expected TD error for a given state-action pair.\footnote{We have dropped the goal subscript ($g$) here, since these concepts do not depend on this assumption.} This is identical to Eq. \ref{eq:goalprogress} except that it only takes the immediate TD error into account, and it is conditioned on action.

The advantage function plays a special role in policy gradient algorithms. Letting $\theta$ denote the parameters of policy $\pi_\theta$, the policy gradient generally takes the following form:
\begin{align}
    \nabla_\theta \mathbb{E}\left[ \sum_{t=0}^\infty \gamma^t r_t \right]  = \mathbb{E}\left[ \sum_{t=0}^\infty \Psi_t \nabla_\theta \log \pi_\theta(a_t|s_t) \right],
\end{align}
where $\Psi_t$ is an unbiased estimate of the value up to a state-dependent baseline. The choice $\Psi_t = A^\pi(s_t,a_t)$, where the baseline corresponds to $V^\pi(s_t)$, achieves the lowest possible variance for an unbiased estimator.

In practice, agents rarely have direct access to the advantage function; instead, they rely on an estimator. This can introduce bias unless some specific conditions are met \citep{sutton2000policy,wen2021characterizing}. The variance of practical advantage estimators can be reduced by taking an average of $N$-step TD errors \citep[\emph{generalized advantage estimation};][]{schulman2015high}:
\begin{align}
    \hat{A}^\pi(s_t,a_t) = \sum_{k=0}^\infty (\gamma \lambda)^k \delta_{t+k},
\end{align}
where $\lambda \in [0,1]$ is a weighting parameter that controls the bias-variance trade-off. When $\lambda = 0$, we recover the standard one-step advantage estimator used in actor-critic methods \citep{barto2020looking}. This estimator has high bias but low variance. When $\lambda = 1$, we recover an estimate of EPE. This estimator is unbiased but potentially has high variance. Intermediate values of $\lambda$ can achieve a balance between bias and variance.

\subsection*{Meta-losses and meta-learning}

A standard machine learning setup starts with a loss function and then derives a learning algorithm for optimizing that loss. An important insight was that both the loss and the learning algorithm could themselves be learned by defining an `outer-loop' that optimizes a meta-loss \citep{zheng2018learning,xu2018meta,xu2020meta,bechtle2021meta,kirsch2020improving}. In order to prevent the loss from becoming vacuous (e.g., by setting every output to have the maximal reward---a form of ``reward hacking''), these approaches typically yoke the meta-loss to some objective measure of task performance (typically through a meta-gradient). Thus, these approaches do not learn truly subjective loss functions. One way to think about the benefit of meta-losses is that they postulate additional \emph{loci of knowledge} beyond the traditional loci of machine learning parameters \citep{zheng2020can}. For example, knowledge about shared structure across tasks can be stored in the parameters of a reward function.

\section{Conclusions}

Where do objective functions come from? This paper proposed an objective-generating subjective function based on expected prediction error. It has some appealing mathematical properties: it is designed to achieve broad and efficient goal coverage, it makes use of existing computational machinery (TD learning), it's hard to game by self-deception, and it's robust to the choice of state representation. It has many connections to earlier ideas (intrinsic reward, learning progress, generalized advantage estimation) and empirical phenomena in psychology and neuroscience (hedonic adaptation, preference for increasing reward, information avoidance/demand, conditional rationality in goal pursuit).

The framework for subjective functions is not yet fully worked out as a practical algorithm. The important questions for future work concern both practical implementation questions as well as questions about the adequacy of expected prediction error as a theory of human goal selection. For example, under what parameter setings does expected prediction error work well (in the sense of goal coverage, efficiency, convergence)? How scalable is the approach to vast goal spaces? These implementation questions can be answered with new computational experiments. The cognitive science questions can be answered with new human experiments: can the theory quantitatively predict goal selection behavior? Can it capture individual variability, for example through variations in the $\alpha$ parameter controlling the balance between prediction error and value?

Another fruitful direction will be to make closer contact with neuroscience. Given the extensive evidence for dopamine signaling of prediction errors, can we show that dopamine has a causal role in goal selection? This will require experimental designs suitable for animal research, in combination with interventions such as optogenetic stimulation.

\subsection*{Acknowledgments}

I'm grateful to Ellie Holton, Ryan Bahlous-Boldi, Pulkit Agrawal, C\'edric Colas, John Vastola, and Kazuki Irie for helpful feedback. This work was supported by the Kempner Institute for the Study of Natural and Artificial Intelligence, a Polymath Award from the Schmidt Sciences, and the Department of Defense
MURI program under ARO grant W911NF-23-1-0277.

\bibliographystyle{apalike}
\bibliography{bib}

\end{document}